\documentclass[a4paper,fleqn]{cas-sc}

\usepackage[authoryear]{natbib}
\usepackage{amsmath}
\usepackage{float}
\def\tsc#1{\csdef{#1}{\textsc{\lowercase{#1}}\xspace}}
\tsc{WGM}
\tsc{QE}
\tsc{EP}
\tsc{PMS}
\tsc{BEC}
\tsc{DE}

\begin{document}
\let\WriteBookmarks\relax
\def\floatpagepagefraction{1}
\def\textpagefraction{.001}

\shorttitle{Learning Spatial Distribution of Long-Term Trackers Scores}

\shortauthors{V. M. Scarrica and A. Staiano}

\title [mode = title]{Learning Spatial Distribution of Long-Term Trackers Scores}

%
\author[1,2]{VM Scarrica}[type=editor,
                        auid=000,bioid=1,
                        prefix=Mr.,
                        role=PhD. Student,
                        orcid=0009-0008-4640-2693]

\cormark[1]

\ead{vincenzomariano.scarrica001@studenti.uniparthenope.it}


\credit{Conceptualization of this study, Methodology, Software}

\affiliation[1]{organization={National PhD Program in AI – Agrifood and Environment, University of Naples Federico II,  (Italy)},
    addressline={Corso Umberto I 40 }, 
    city={Naples},
    postcode={80138}, 
    country={Italy}} 

\author[2]{A Staiano}[%
   prefix=Prof.,
   role=Co-ordinator
   ]

\ead{antonino.staiano@uniparthenope.it}

\credit{Data curation, Writing - Original draft preparation}

\affiliation[2]{organization={University of Naples Parthenope, Department of Science and Technology},
    addressline={Centro Direzionale Isola C4}, 
    city={Naples},
    postcode={80143}, 
    country={Italy}}

\cortext[cor1]{Corresponding author}

\begin{abstract}
Long-Term tracking is a hot topic in Computer Vision. In this context, competitive models are presented every year, showing a constant growth rate in performances, mainly measured in standardized protocols as Visual Object Tracking (VOT) and Object Tracking Benchmark (OTB). Fusion-trackers strategy has been applied over last few years for overcoming the known re-detection problem, turning out to be an important breakthrough. Following this approach, this work aims to generalize the fusion concept to an arbitrary number of trackers used as baseline trackers in the pipeline, leveraging a learning phase to better understand how outcomes correlate with each other, even when no target is present. A model and data independence conjecture will be evidenced in the manuscript, yielding a recall of 0.738 on LTB-50 dataset when learning from VOT-LT2022, and 0.619 by reversing the two datasets. In both cases, results are strongly competitive with state-of-the-art and recall turns out to be the first on the podium.
\end{abstract}


\begin{highlights}
\item Generalization on Tracker-fusion strategy
\item Model and data long-term independence 
\item Introduction of a new out of view class
\item First recall achieved on two distinct benchmarks
\item Extension of the paradigm to multi-object tracking
\end{highlights}


\maketitle

\section{Introduction}

In this section the main topic will be introduced, namely long-term tracking and the context in which this task, and more generally object tracking, is defined. In addition, some of the most well-known issues concerning it will be described. 

\subsection{Definitions and taxonomy}
Object Tracking has caught the attention of most Computer Vision experts in recent years. A less formal definition of it could be, given in input a sequence $S={f_{1}...f_{K}}$, with a set of groundtruth targets $T={g_{1}...g_{K} | g_{i} \in f_{i}} \forall i \in [1,K]$, look for the set of predictions $P={p_{1}...p_{K} | p_{i} \in f_{i}} \forall i \in [1,K]$ such as to minimize the differences of each prediction, respectively, from each target. The first occurrence is given from the groundtruth and it consists of the chosen object, which will be used as a template by the tracker.
This definition, in fact, can be adapted depending on the measurement context in which the task is to be included. \textit{Short-Term Tracking}, for example, concerns sequences in which the target does not disappear from the scene, and on which therefore \textit{re-set} can be carried out in a lawful way, where re-set means a reinitialization of the target. Different is the speech of the \textit{Long-Term}, where the target can leave the scene in whole or in part, but no re-set allowed. If the tracker predicts incorrectly with respect to the target or predicts an object when the target is not present, it is called \textit{tracking failure}. The choice of the initial target is also a very important parameter for the execution of a long-term tracker (and a tracker in general): the more detailed and less noisy the template will be compared to the background, the more robust and consistent there will be in the results along the sequence. An important distinction must be made between \textit{online} trackers and \textit{offline} trackers. The former exploit only the knowledge of the current frame, while the latter also incorporate information from other frames in the sequence, to improve the reconstruction of the path to be traced. Extending the discourse to more objects, the same conditions can apply to each individual target to be followed. In this case we speak of \textit{Multi-Object Tracking}. The two main differences between single and multi object approaches concern first the choice between favoring the accuracy of the single object or the efficiency in tracking at the expense of individual predictions, and then the number of classes considered if an object classification phase is applied further to the object detection already used by default by the tracker. As for the first aspect, at the state of the art multi-object trackers are more efficient but less accurate on individual predictions, and vice versa single-object trackers. On the second aspect, single-objects are often generalized to multi-class paradigms, while multi-object trackers tend to focus more on single-class tasks to simplify their complex pipeline, e.g., human and crowd tracking (\citet{shao2018}). FairMOT (\citet{zhang2020}) is one of the most interesting state-of-the-art models from the point of view of comparisons, as it is often extremely competitive and generalizable to multiple classes through specific fine-tuning phases. Usually, most trackers designed to date use predictions in the form of bounding boxes, but there are works that use instance segmentation-based phases in their architectures to improve final predictions, such as RTS (\citet{paul2022}). Opening the discussion to multiple dimensions, for 3D images reconstructed by stereoscopy or with multiple calibrated cameras (\citet{Ming2019}), there are interesting 3D-object trackers methods in the literature, \citet{hu2021}. Long-Term Tracking can be further divided into \textit{Re-Detection Long-Term}, where in accordance to a visibility confidence score the tracker can arbitrarily choose to re-detect the target, or \textit{Pseudo-Long-Term}, where the target is never re-detected. In this work, Long-Term Fusion-Trackers strategies will be discussed, and a conjecture on a data and model independent learning procedure on a generalized number of Long-Term Trackers scores will be made for boosting performances. 

\subsection{Environment and issues}
Object Tracking involves various areas of study in science, and below will be mentioned some of current importance. \textit{Human Interaction} can certainly be mentioned (\citet{singh2019}), for example in the recognition and tracking of gestures taken by a web-cam for subsequent processing into commands to be executed by the machine. Object tracking also finds applications in \textit{intelligent monitoring} (\citet{Tai2004}), for example on work sites where workers must be monitored and tracked for safety and control reasons. \textit{Automated driving} also uses Object Tracking techniques to monitor the trajectories of pedestrians and vehicles around (\citet{tang2019}), so as to avoid collisions. In \textit{virtual reality}, objects can be tracked to reproduce effects on them. Object Tracking can also be used in \textit{Surgical Navigation} to follow remote interventions, and impart movements of particular tools, such as a scalpel, to a robot. In the forensic field, Object Tracking is used for \textit{Crime Prediction} in video surveillance systems (\citet{Miao2016}), where subjects can commit offenses or access places without authorization, or at unauthorized times. In the military, where the deployment of smart weapons is increasingly required, Object Tracking is the second candidate technology after IR for \textit{Navigation and Reconnaissance} (\citet{lei2015}), for example missile warheads.
As in all branches of Computer Vision, where techniques to solve difficult problems have limitations due to conditions of infeasibility in visual input, even the tracking of objects has some fundamental problems that it is necessary to argue, as already counted by considerable publications like \citet{wu2013}. For example, the \textit{scale variation} introduces an important difficulty related to the change of perspective in 2D, and the change of resolution of the template may have to require more effort on the part of the tracker in extracting features invariant to the scale. The \textit{lighting conditions} are another fundamental element: the reflection of light, backscatter, diffusion, refraction and other phenomena of visible waves hinder a good success. Sequences can also have \textit{occlusions}, which are areas where the target is partially obscured by other objects or partially disappears from the scene. The objects to be tracked can also undergo similar and non-similar transformations, then deform in order to fool the tracker. An important feature must therefore be robustness to \textit{morphological changes}. Other transformations can be \textit{rotations} on various planes and reference axes, \textit{blurring} and \textit{resolution reduction}, as well as \textit{noise} introduced by the hardware used or by the filtering software in post-processing. The object can be confused with similar objects belonging to the background, and in this case we speak of \textit{background clutter}. This turns out to be an extremely complex issue that is much debated in the most important competitions. The framerate turns out to be another important parameter, because if an object moves very fast, then in \textit{fast motion}, you can lose information and therefore also lose the template. The displacement between two frames in terms of ground clearance would be too wide. Finally, the most complex problem is probably the \textit{out of view} (OoV), where the object can actually disappear from the scene and then return to it, even in different shapes and colors. This is the main problem that must be addressed, together with the others already envisaged, Long-Term Tracking.

\section{Related works}
This section will serve as a survey on the most used object tracking techniques in the history of Computer Vision, up to the latest algorithms, with a focus on Long-Term Tracking and merger approaches.

\subsection{Image Processing}
Starting with methodologies based on image processing, which today would be rudimentary, we must necessarily remember the first searches on the target set, which took place by searching for the same target (defined as template), within a region of interest (ROI). The template, defined in its instance, is called a patch, and initially scrolled along both dimensions of the current frame, considering the entire window as ROI. The way the template was defined was variable. We started by considering the same levels of pixel intensity, then wanting to make a real template matching, resulting in very poor long-term evaluation performance; After that, it was considered appropriate to vary the patch by reinitializing it every $n$ frames, when the matching score was below a threshold, but even this limited the tracker's ability to analyze long sequences. From simple template matching, we have moved on to the consideration of more elaborate statistical measures, such as correlation (\citet{Xing2021}). 
\begin{equation}
G(i,j) = \sum_{u=-k}^{k} \sum_{v=-k}^{k} F(u,v)I(i+u,j+v)
\label{eqn1}
\end{equation}
From the eqn. \ref{eqn1}, you can see how the correlation takes frame I and kernel K as input, with kernel size strictly smaller than frame size.
In the same way as template matching, these measures could be maintained from the initial template throughout the sequence or updated during construction by reinitialization. However, the search on the entire window can generate an abnormal number of false positives, also due to the presence of objects similar to the target, so the size of the ROI has been drastically lowered to a local neighborhood, called search window.
\begin{equation}
SW(i,j) =\{I(i,j)\,\mid\,|I(i,j) - SW(i,j)| < \delta, \forall i,j \in I\}
\label{eqn2}
\end{equation}
In eqn. \ref{eqn2} there is a definition search windows, where $\delta$ is a positive threshold with value smaller than the entire window size. This threshold can also be adaptive, depending on the algorithm one chooses. Correlation filter-based methods are however subject to sensitivity to transformations such as rotations, morphological changes and sudden changes of direction in the trajectory. However, one of their advantages remains translational invariance, under certain steady-state assumptions. Moving on to motion estimation-based approaches, perhaps the most used method to solve the tracking problem is the Kalman filter (\citet{ali1998}). It consists of a probabilistic model that, based on an a priori mean state $x_{k|k-1}$ and a priori covariance $p_{k|k-1}$, approximates the predictions of the state a posteriori and the error of a posteriori covariance, through a step of updates to the parameters of the model:
\begin{equation}
x_{k|k} = \theta x_{k|k-1} + \rho
\newline
p_{k|k} = \theta p_{k|k-1} \theta^T + \psi
\label{eqn3}
\end{equation}
In the eqn. \ref{eqn3}, both state variables are updated through optimization of parameters $\theta, \rho$ and $\psi$. Its limitations consist in considering stationary linear dynamical systems, whereby sudden changes in the direction of the object would be predicted with a big error on covariance. However, the Kalman filter is found in many other algorithms, grafted as a more complex piece of pipeline. In addition to methods based on motion estimation, image processing has produced other systems to try to follow objects in videos, such as those based on histograms. By calculating the histogram of a patch, one can approximate a probability density function of pixel intensity levels, and set it as a similarity criterion. Known algorithms that use histograms are the MeanShift (\citet{ali1998}), and its adaptive evolution, the CAMShift (\citet{Bradski1998ComputerVF}). Despite their revolution in the field of object tracking, it is notorious that histograms do not capture topological information, which is of fundamental importance when the targets to be followed have detailed textures, since they do not detect occlusions well or can be confused with similar objects. Compared to previous methods, however, they can detect morphological changes.

\subsection{Machine Learning}
Machine Learning (ML) techniques have practically completely replaced most solutions based on image processing in Computer Vision, both for their efficiency and for their accuracy, sometimes stable and sometimes even better than the most rudimentary techniques. As for classical Machine Learning, where a classifier or regressor such as Random Forest (RF, \citet{breiman2001}), K Nearest Neighbors (K-NN,  \citet{cover1967}) or Support Vectors Machine (SVM, \citet{cortes1995}) is used to learn features extracted from a patch, we can mention the work of \citet{tian2007}, where an Ensemble of SVMs is used to trace objects; on the contrary, \citet{thormann2017}, describe a first RF-based system to learn the results deriving from a Computer Vision algorithm to try to replace the latter. This last work can be considered as a source of inspiration for the study carried out in this paper, although it has different conceptual bases. More performing and commonly used to approximate object trackers, are the Deep Learning Object Detection algorithms based on convolutional networks, including YOLO (\citet{redmon2016}), Faster-RCNN (\citet{ren2015}). Mainly, these Object Detectors are used to search for the target object in each frame by training only on the patch of the first frame of the sequence, and if possible, updating their weights according to their own predictions. Their main disadvantages are inductive bias and anchor dependence.
Multi-Domain Convolutional Neural Network (MDNet, \citet{nam2016}) is composed of a structure of convolutional layers followed by a series of parallel branches, each representing a different domain, where in the case of tracking a sequence is assumed as a domain. First, each branch trains on the single sequence, after which the shared convolutional layers are trained to give the model global knowledge. The type of classification is binary, that is, to distinguish foreground and background. Different is the approach of Siamese networks, in which two models having the same parameters are put in parallel, and they are given as input the entire frame and the patch to be searched. After their execution, an aggregation function is applied (cross-correlation is widely used) to obtain the final heatmap where, through an appropriate rescaling, the result will be displayed. Known Siamese network models are Siam-RCNN (\citet{voigtlaender2019}), Siam-FC++ (\citet{xu2020}), Siam-RPN++ (\citet{li2018}), Siam-Mask (\citet{wang2019}). This approach was considered the state of the art until a few years ago, then supplanted by the introduction of transformers. The main defects of Siamese networks concern the poor ability to learn the background as a function of the foreground (and therefore strengthen its discrimination) and the lack of reliability of the output score, which unlike other probabilistic models, turns out to be an index of similarity. Later, transformers-based solutions began to take hold due to their excellent ability to learn sequences both in a spatial and temporal sense. Transformers can in fact be compared to Recurrent Neural Networks (RNNs) that are much less expensive in terms of training performance, but require large amounts of sequences. As the RNNs do, they take as input a sequence and return a sequence, but they introduce a new data relation study function called \textit{attention}. These relationships are studied by encoding incoming data once a maximum token size is established. Positional encoding are systems that allow you to rearrange the input in order to simplify the calculation of attention. Transformers usually appear as auto-encoder structures, and their training includes the aid of three fundamental matrices: $Q$, $V$ and $K$, or, respectively, \textit{queries}, \textit{values} and \textit{keys} (\citet{vaswani2017}). As in a retrieval system, the query can be considered the search string, the keys the domain in which to search and the values the final result. 
\begin{equation}
c_i = \sum_{j}^{} a_ij h_j \newline\textbf{where}\newline  \sum_{j}^{} a_j = 1
\label{eqn4}
\end{equation}
In a first version of Transformers definition (\citet{bahdanau2016}), the attention was defined as in eqn. \ref{eqn4}, where $h_j$ are the values and $a_j$ the coefficients to be pursued. (\citet{bahdanau2016}) proposed a neural network for learning these scores.
The calculation in this case turned out to be too expensive, since it maps directly from a sequence of dimension N for the encoder to a sequence of size M for the decoder. 
\begin{equation}
 Attention(Q,K,V) = softmax(\frac{QK^T}{\sqrt{d_k}})V
\label{eqn5}
\end{equation}
By choosing to project the sequences on a common space (\citet{vaswani2017}), through the function $f(x)$ for the encoder and $g(y)$ for the decoder, we obtain projection vectors called keys $K$ for the encoder and queries $Q$ for the decoder. In eqn. \ref{eqn5}, an evolved definition of attention is given, with $d_k$ the queries and keys dimension.
\begin{equation}
MultiHead(Q,K,V) = Concat(head_1...head_h)W^0 \newline\textbf{where}\newline head_i = Attention(QW_i,KW_i,VW_i) 
\label{eqn6}
\end{equation}
In eqn. \ref{eqn6} transformer layers are described, often multi-head. When the queries, values and keys are inputted from the same sequence, we talk about \textit{self-attention}; if the queries come from a sequence, while keys and values are from another sequence, we deal with \textit{cross-attention}. Usually, the first approach is used for unsupervised language models, like GPT-3, \citet{brown2020}. The latter concern models like Stable Diffusion (\citet{rombach2022}), where more than an image, there is a prompt text input for generating text-driven manipulated images.
Mentioning some of their main bottlenecks, transformers suffer of poor explainability in their architecture and they need huge amount of data to be trained onto, that makes them very difficult to be run on medium power machines. 
In Short-Term universe, There are several transformers used to solve the tracking problem. Among the best known must be counted STARK (\citet{yan2021}), SuperDiMP (\citet{bhat2019}), KeepTrack (\citet{mayer2021}) and VitTrack (\textit{Chen et al.} in \citet{kristan2023}). STARK (\citet{yan2021}) follows a classic auto-encoder structure, with a backbone for feature extraction and a branch for bounding box prediction. It stores an initial template and a dynamic template. In order to make the prediction, first check through a score head if it is necessary to change the dynamic template, and if so replace it with the previous one, updating it. SuperDiMP (\citet{bhat2019}) follows a different architecture, applying, after the extraction of the training features, i.e. the stored templates, a model prediction branch. This step consists in calculating the weights of a convolutional model based on an initialized model and an optimization algorithm that iteratively adapts the former to the training distribution. After that, the predicted model is used on the features extracted from the test frames, in order to produce the final score map. KeepTrack (\citet{mayer2021}) uses SuperDiMP within its pipeline as a baseline tracker on two consecutive frames. Then, having obtained two candidate targets with relative score maps, a feature encoding is made to the two maps. Defined as a Graph Neural Network (GNN) to which the embedding candidates are passed, the associations between the map elements are calculated and learned, seen as nodes connected by self-attentional and cross-attentional edges. Then to perform the candidate matching and obtain by exclusion the final target, a Sinkhorn based algorithm is used. ViTTrack (\textit{Chen et al.} in \citet{kristan2023}) is based on the ViT transformer (\citet{dosovitskiy2021}). To this, he adds a corner prediction head. Similar to STARK, the initial patch and template are chained together and used for feature extraction and corner prediction. 

\subsection{Trackers Fusion Strategy}
Throughout the history of object tracking, various alternative ways to simplify processes have been experimented with, which gradually became more and more complex. The very design of a transformer and its training involves in-depth study and huge amounts of data.
To get an idea of the amount of data that would be needed to describe an associative memory capable of solving the problem of object tracking, consider that a frame has dimensions $W$ and $H$ in $2D$, and each pixel can take $N$ values. The possible frames, assumed as points in an unbounded discrete space $W \times H$ dimensional, are $N^{W \times H}$. This value alone represents the set of possible groundtruths in an image classification problem. Introducing the third dimension for sequences, we say that every sequence has a length equal to $K$, with $K$ not variable, and every frame in each sequence is different from every frame from other sequences. Assuming that an initial target is composed of a subset of pixels of the first frame, the amount of subsets is $2^{W \times H}$; assuming even that for each initial target, there is an unique and distinct series of groundtruth subsets for the entire sequence, and among the sequences, below we can observe a calculation of a lower bound of possible groundtruths necessary to construct an associative memory for object tracking.
\begin{equation}
OTG = \frac{N^{W \times H} \times 2^{W \times H}}{K} = \frac{2N^{W \times H}}{K}
\label{eqn7}
\end{equation}
In the eqn. \ref{eqn6}, $OTG$ stands for object tracking groundtruths and the unit of storaging is in bytes. Wanting to go into the practical field, assume that a video has a resolution of $1280 \times 720$, with frame rate of 30 fps, length of 60 seconds and an encoding of 3 bytes for each pixel; Using the eqn. \ref{eqn6}, you get a value of $8.65 \times 10^{717128} $ terabytes of data. An average algorithm based on deep learning can be used on at most a quantity of the order of terabyte units, and this is already expensive. To reduce the complexity of these algorithms, it was decided to merge multiple trackers into a parallel pipeline, and to aggregate the results into a single branch. Going backwards, (\citet{vojir2016}) uses an adaptive Hidden Markov Model to predict which tracker to use among a pool of complementary trackers; Falling within the field of deep learning, in Long-Term tracking, until 2020 the approaches considered were sequential and the pipelines were linear. After predicting the bounding box and confidence score, the tracker decided whether to re-detect the target based on a threshold or learning. An example of tracker that uses this approach and that has obtained the best F1 scores on the VOT-LT2019 (\citet{kristan2019}) and VOT-LT2020 (\citet{kristan2020}) challenges is LT\_DSE (\citet{kristan2019}) (winner of both editions). Since 2020, the Tracker fusion strategy paradigm has also begun to be adopted in Long-Term tracking and using deep learning methods. In particular, the winner of the VOT-LT2021 (\citet{Kristan2021}) challenge was mlpLT (\citet{dunnhofer2022}), based on the merger of the STARK and SuperDiMP trackers, with an online verification phase by MDNet. This tracker also applies the so-called correction phase, i.e. the tracker that is evaluated as better between the two will give a result that will act as a template for both trackers. An improved version of its F1 score is CoCoLoT (\textit{Dunnhofer et al.} in \citet{kristan2023}), which replaces the SuperDiMP baseline tracker with KeepTrack. The VOT-LT2022 (\citet{kristan2023}) challenge was, instead, won by VITKT\_M (\textit{Zhang et al.} in \citet{kristan2023}). This model consists of a composition of the ViTTrack and KeepTrack trackers, followed by the metric model MetricNet (\citet{zhao2020}). It is then extended by adding a motion module that predicts the trajectory of the current target when it assumes abnormal behavior.
The present work does not only want to propose a series of innovative models, but wants to act as a standard of generalization for tracking fusion strategy. In particular, parameterize the number of baseline trackers, introduce the classification of an OoV state, not considered by most works. The latter is of fundamental importance, especially in military and medical tasks, where the number of false positives and true negatives must be minimised. In addition, as will be explained in the \textbf{Methodology and materials} section, the process of choosing the tracker result to rely on for the final result will be treated as a learning procedure acted by a generic learner. These features will infuse the model with the ability to abstract itself from the type of algorithm used, both from the point of view of baseline trackers and learners. As will then be shown by the experiments, the models will also have the ability to abstract from data: using LTB-50 (\citet{lukezic2021}) as a training dataset and VOT-LT2022 evaluation dataset as a test set, the recall obtained will be the highest ever, with a highly competitive F1 score; similarly, using VOT-LT2022 as a training set and LTB-50 as a test, the results will remain almost unchanged. In the experiments, ablations and modifications will be introduced to avoid any kind of a priori knowledge on the test set.

\section{Methodology and Materials}
In this section the most used protocols in the context of Long-Term Object Tracking will be described, the benchmarks that are used by the model both as training sets and as test sets, a functional representation of how $N$ trackers can be complementary to each other and finally the model itself.
\subsection{Evaluation protocols}
Nearly all tasks in Computer Vision gain international visibility not only because of the inherent complexity that lies in their problems, but because of the way in which the type of solutions proposed to them is evaluated. The Object Tracking Benchmarks (OTB) (\citet{wu2015}) and Visual Object Tracking (VOT) (\citet{Kristan2014}) protocols are de facto standards in the field, and most state-of-the-art models refer to them. To better define their types of evaluation, it is necessary to introduce the concept of Intersection over Union (IoU) between two bounding boxes or between two masks.
\begin{equation}
IoU = \frac{| r_t \cap r_p |}{| r_t \cup r_p |}
\label{eqn8}
\end{equation}
In eqn. \ref{eqn8} the $r_t$ stands for the target ragion while the $r_p$ for the predicted. IoU is widely used for tasksof all kinds, from instance segmentation, to semantic segmentation, and it is comprised in an interval between 0 and 1. In addition, the definition of Average Center Location (ACL) is needed.
\begin{equation}
ACL = E(||x-y||_2) 
\label{eqn9}
\end{equation}
In eqn. \ref{eqn9} for $x$ the target bounding box or mask center is intended, while for $y$ we also mean that for prediction. The OTB protocol is known to be a one-step protocol, that is, launched and never stopped on a specific frame of the sequence, even on the short-term in the presence of tracking failure. In practice, it never allows resets. Its main metrics are \textit{accuracy} and \textit{robustness}. Given two thresholds $\lambda$ and $\delta$, in accuracy we can distinguish three measures \textit{precision}, \textit{success} and \textit{Area Under Curve} (AUC).
\begin{equation}
P = \%f_i \forall i \in [1,K] | ACL < \lambda 
\label{eqn10}
\end{equation}
\begin{equation}
S = \%f_i \forall i \in [1,K] | IoU > \delta
\label{eqn11}
\end{equation}
\begin{equation}
AUC = \int_\delta{IoU}~\textbf{with}~\delta \in [0;1]
\label{eqn12}
\end{equation}
In the eqn. \ref{eqn10} and eqn. \ref{eqn11} $\%f_i$ indicates the number of frames belonging to the sequence set $S$, in the first case having threshold on the ACL with respect to $\lambda$, in the second case on the IoU with respect to $\delta$. The AUC (eqn. \ref{eqn12}) is instead the integration of the IoU with respect to the change in the $\delta$ threshold in the range $[0;1]$.
As for the calculation of robustness, it refers to evaluating accuracy in three different ways: One Pass Evaluation (OPE), Temporal Robustness Evaluation (TRE) and Spatial Robustness Evaluation (SRE). OPE refers to run the evaluation in one step with no reset, TRE divides the sequence into segments and executes OPE on each individual segment, and then mediates the results; finally, SRE applies OPE on 12 transformations of the same sequence, based on augmentation.
As for the VOT protocol, it identifies a different evaluation criterion for each type of tracking: for the Long-Term the metrics considered are precision, recall and F1-score, calculated according to a threshold linked to IoU.
\begin{equation}
\tau_{\sigma} = max\{\tau|max_{\tau} F(\tau)\} 
\label{eqn13}
\end{equation}
\begin{equation}
Pr(\tau_{\sigma}) = \int_{0}^{1}{Pr(\tau_{\sigma},\tau_{\Omega})d_{\tau_{\Omega}}} = \frac{1}{N_p}\sum_{t\in \{t:G_t \neq \emptyset\}}\Omega(A_t(\tau_{\sigma}),G_t)
\label{eqn14}
\end{equation}
\begin{equation}
Re(\tau_{\sigma}) = \int_{0}^{1}{Re(\tau_{\sigma},\tau_{\Omega})d_{\tau_{\Omega}}} = \frac{1}{N_g}\sum_{t\in \{t:A_t(\tau_{\sigma}) \neq \emptyset\}}\Omega(A_t(\tau_{\sigma}),G_t) 
\label{eqn15}
\end{equation}
\begin{equation}
F(\tau_{\sigma}) = \frac{2Pr(\tau_{\sigma})Re(\tau_{\sigma})}{Pr(\tau_{\sigma})+Re(\tau_{\sigma})}
\label{eqn16}
\end{equation}
As can be seen from \ref{eqn13}, the calculation of the three metrics is carried out by searching for the threshold $\tau_{\sigma}$ that maximizes the F1-score.
Precision and recall (eqn. \ref{eqn14} and eqn. \ref{eqn15}, respectively) integrate their respective measures in variation at the threshold of IoU in the range $[0;1]$, where $A_t$ stands for predicted bounding box, $G_t$ stands for groundtruth and $\Omega$ indicates the overlap operator. The situation is different for the F1-score (eqn. \ref{eqn16}), where the maximized threshold in eqn. \ref{eqn13} is taken into account to give the final outcome.
There is therefore no pre-set threshold value to refer to. The latter protocol was used in the experiments, in accordance with the results presented at the last reference challenge, namely the VOT-LT2022 (\citet{kristan2023}).
\begin{figure}[ht]
	\centering
        \includegraphics[scale=.43]{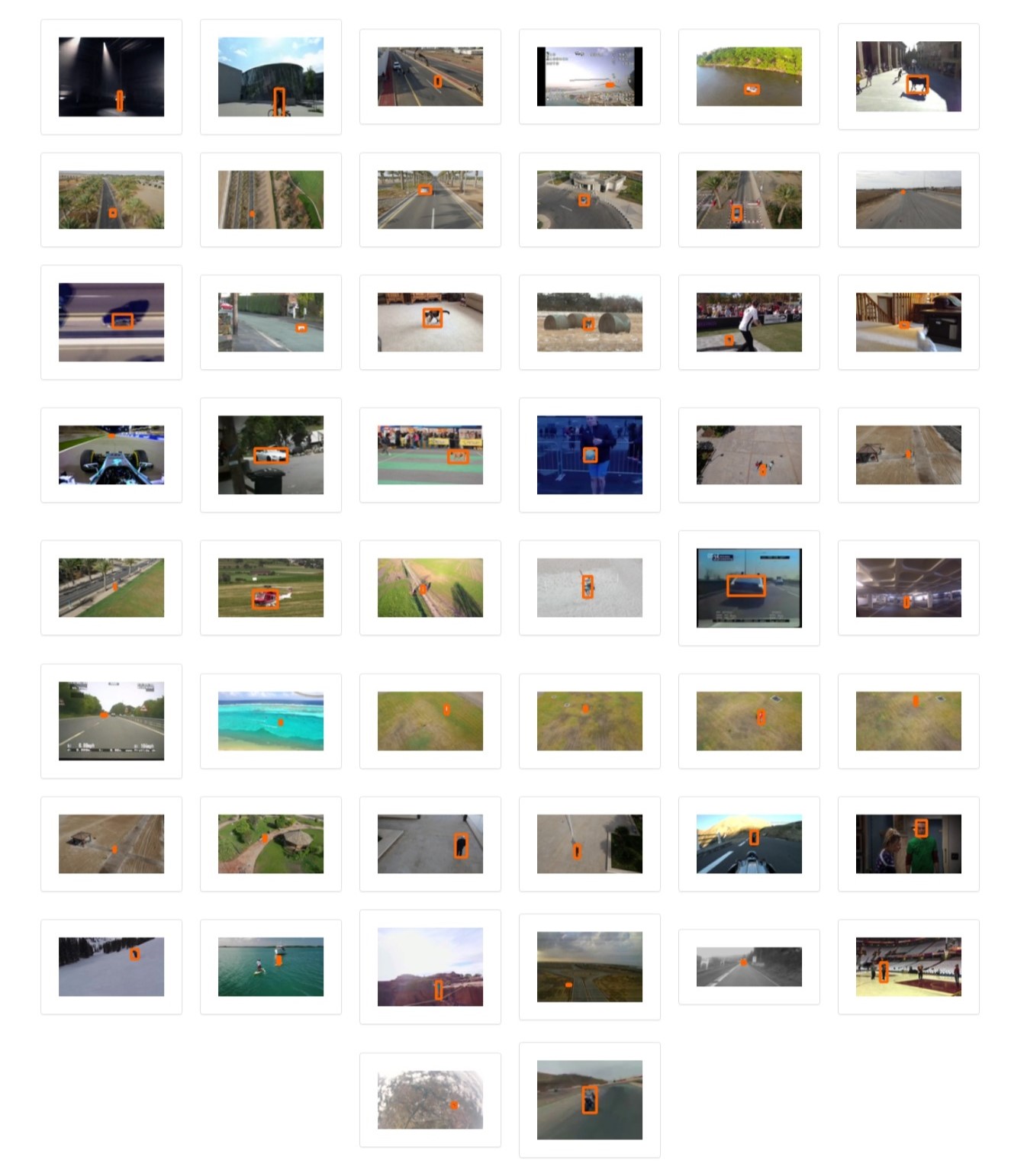}
	\caption{In the illustration are visible the 50 sequences of LTB-50, among which there are some of the most complex issues covered in the introduction, such as out of views or transformations of various types. The cut has been made on \url{https://www.votchallenge.net/vot2019/dataset.html}}
	\label{ltb50}
\end{figure}
\begin{figure}[ht]
	\centering
		\includegraphics[scale=.43]{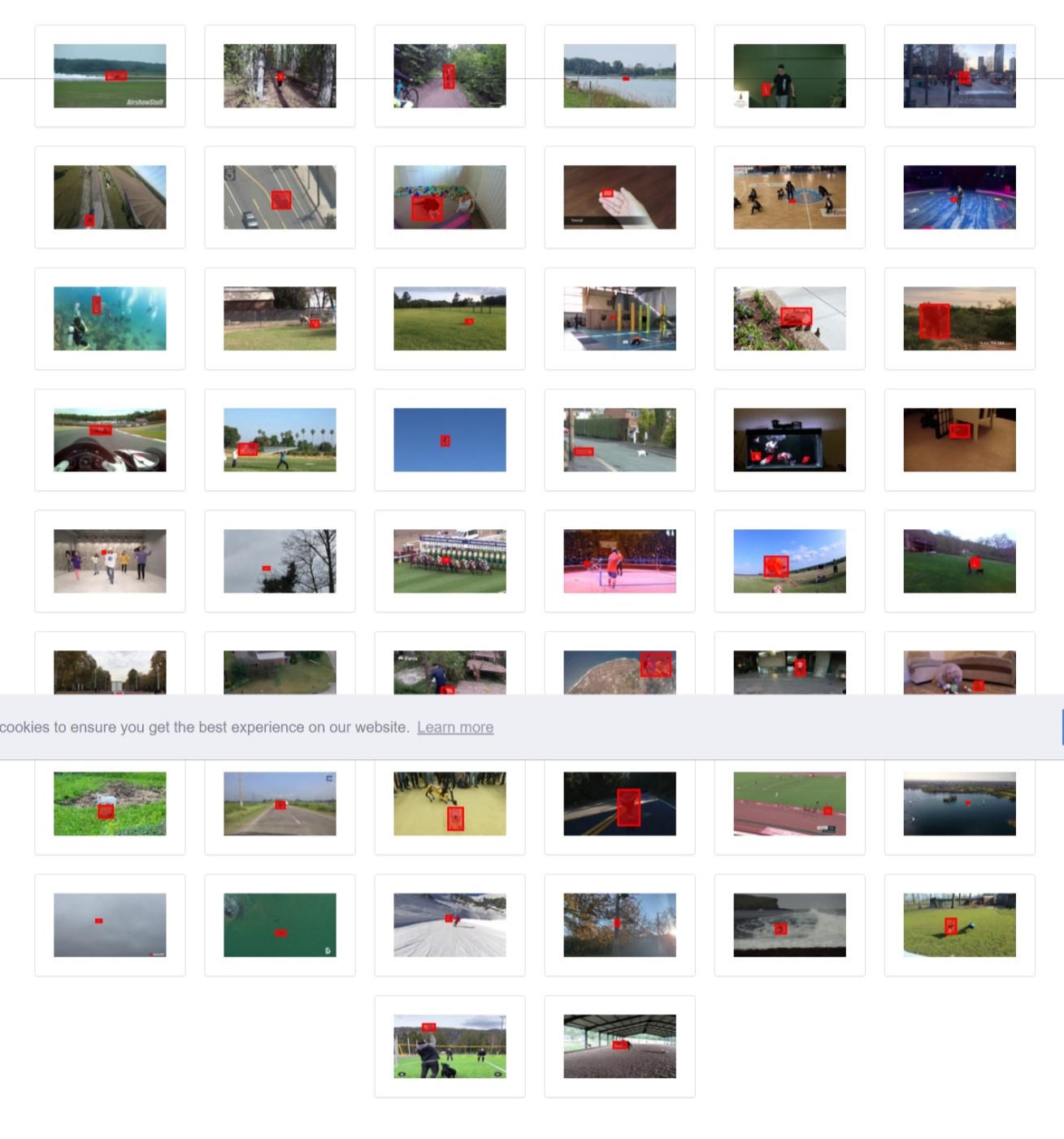}
	\caption{In the illustration are visible the 50 sequences of VOT-LT2022, which are found to have within them a smaller number of frames overall, but a higher resolution. The cut has been made on \url{https://www.votchallenge.net/vot2022/dataset.html}}
	\label{vot2022}
\end{figure}
\subsection{Benchmarks}
As for the details on the datasets used in the pre-training phase by the models that have been mentioned and that have been used in the proposed work, these can be found directly on the individual reference papers. The datasets that will be referred to in the manuscript and that have actually been used in the experiments, both for training and testing, are LTB-50 (\citet{lukezic2021}, adopted for the VOT-LT 2019, 2020, 2021 challenges) and VOT-LT2022 (\citet{kristan2023}).

The LTB-50 dataset is composed of 50 sequences, for a total of 215294 frames, divided unequally between the sequences. The sequences have different resolutions and a wide variety of target subjects, \textit{e.g.}, animals, people, cars, etc. It contains within its scenes most of the problems dealt with in the Environment and issues section, such as OoVs, geometric transformations, different visible wave phenomena, poor acquisition quality. Implicitly, despite having constant frame rates, subjects move at different speeds, presenting rates of fast motion varying between sequences. In the experiments, the dataset will be used both in the training phase and in the test phase, exploiting the annotations produced by the VOT community during the creation phase of the benchmark. A graphic testimony of its scenes can be found in Fig. \ref{ltb50}.

Similarly to LTB-50, the VOT-LT 2022 dataset, introduced only for the 2022 edition (the VOT challenge has in fact changed the tracking task from 2023, bringing it to multi-object tracking) contains 50 sequences, for a total of 168282 frames. They have different resolutions, but the same frame rate. The considerations regarding the issues and the type of targets apply in the same way as the LTB-50 dataset, but it has been empirically noted that VOT-LT 2022 is more difficult to evaluate, as reported by the latest results. On average, the resolution is higher than the LTB-50 dataset, and the dataset is heavier in terms of storaging memory. In the experiments it will be used both as training and as a test. The ensemble of the sequences is shown in Fig. \ref{vot2022}.

\subsection{Trackers complementarity}

Before presenting the spatial learning model, it is necessary to understand what are the theoretical foundations on which the fusion strategy is based. What assumes mathematical validity in merging multiple trackers together is their \textit{complementarity}. Define complementarity as the ability of $N$ generic trackers to return qualitatively different results in a complementary way. Clearly, in the practical field, among them the trackers have different performances, and it is not said that there are situations of fairness in which everyone can, in turn, give their own result without overlapping the others. In this sense, various types of situations can be defined, representable    mathematically. Appealing to the returned outputs, namely the confidence score $c_{ij}$ and the bounding box $b_{ij}$, where $i$ is the frame index in the sequence $S$ and $j$ the tracker index in the ensemble. If and only if, the $IoU_{ij}$ calculated between the $b_{ij}$ and the $g_i$, i.e., the corresponding groundtruth of the $i$-th frame, is better than the other $IoU_{ik, k \neq j}$, then the prediction of the $j$-th tracker is assigned as the corresponding prediction. Note that the comparison is independent of the confidence value, as it is not said that if $c_{ij}$ is greater than $c_{ik, k \neq j}$ then $IoU_{ij}$ is greater than $IoU_{ik, k \neq j}$. In this way, it is obtained that to have the best tracking system acting on a sequence (with the given trackers), it is necessary to subject all its frames to the previous comparison, and have a one-by-one association between a frame $i$ and a tracker $j$ (i.e., the best, or the one whose the prediction should be chosen). For simplicity, we will introduce four scenarios, which will act as extreme conditions in which a multi-tracker system can be presented: \textit{in-phase}, \textit{anti-phase}, \textit{Dirac delta} and \textit{upper limited}. 

\begin{figure}[ht]
	\centering
		\includegraphics[scale=.75]{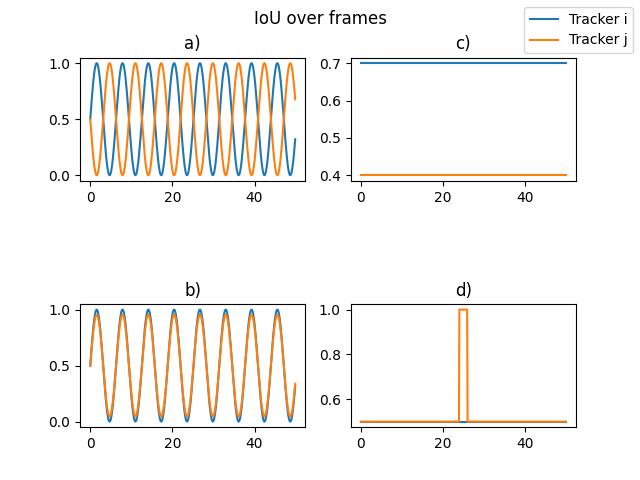}
	\caption{The four extreme scenarios of a multi-tracker system: a) anti-phase trackers; b) in-phase trackers; c) upper limited trackers; d) Dirac delta like distribution.}
	\label{complementarity}
\end{figure}

As can be seen from Fig. \ref{complementarity}, where a generic pair of trackers $i$ and $j$ taken from the system's trackers pool, the four configurations are:
\begin{itemize}
    \item \textbf{Anti-phase}: for each frame of the sequence there is always a tracker with a IoU higher than the others. Trackers alternate their predictions in a round-robin pattern to optimize decision-making. If $N$ trackers preserve this property on a generic sequence, you will always be able to get the maximum performance from their simultaneous execution.
     \begin{equation}
    IoU_j(t) = A_jsin(2\pi ft+\rho_j) 
    \label{eqn17}
    \end{equation}
    In eqn. \ref{eqn17} the IoU function defined on frames domain is reported as a sinusoidal wave, where $j$ stands for the tracker index in the multi-tracker system. Every sinusoid has its own phase $\rho_j$, that makes the round-robin scheme appliable.
    \item \textbf{In-phase}: all the trackers in the system behave in the same way. They admit the same peaks in amplitude on the IoU and therefore the maximum obtainable from the union of their performances is equal to the performance of the individual.
     \begin{equation}
    IoU_j(t) = A_jsin(2\pi ft)
    \label{eqn18}
    \end{equation}
    In eqn. \ref{eqn18} a specific case of eqn. \ref{eqn17} is considered, with $\rho_j=0 \quad \forall j$. 
    \item \textbf{Upper limited}: there is always one and only one tracker (or a subset of the entire pool) that overpowers the performance of the others, thus making the execution of the poorest trackers useless. The maximum obtainable in this configuration is therefore given by the best or by the simultaneous execution of the best.
     \begin{equation}
    IoU_j(t) = K_j
    \label{eqn19}
    \end{equation}
    In eqn. \ref{eqn19} every $j$-th tracker IoU discrete curve is represented by a constant function $K_j$, with $K_j \neq K_i \quad \forall \quad i \neq j$.
    \item \textbf{Dirac delta}: trackers behave in the same way as in the in-phase case, the difference lies in the fact that in a single frame the IoU of one of them (or a subset of the pool) turns out to be greater than the others. This generates a paradoxical situation in which although statistically trackers can be considered equal, one of them or a subset of them can falsely be considered better, as in the case of Upper limited.
    \begin{equation}
    \begin{cases}
    IoU_j(t) = K_j \\
    IoU_i(t) = K_j \quad \textbf{if} \quad  t \neq t_0 \\ 
    IoU_i(t) = K_i > K_j \quad \textbf{if} \quad t == t_0
    \end{cases}
    \label{eqn20}
    \end{equation}
    In eqn. \ref{eqn20} a special case of eqn. \ref{eqn19} is presented: only one point of the best constant function is higher than the others, assuming a Dirac Delta shape.
\end{itemize}

The situations described are clearly ideal and almost impossible to replicate in practice, but they serve to understand how the task of learning the behavior curve among the various trackers is a fundamental task, once it is established that the trackers present among them at least one point of complementarity. This can be verified by a specially chosen training set. 

\begin{figure}[ht]
	\centering
		\includegraphics[scale=.50]{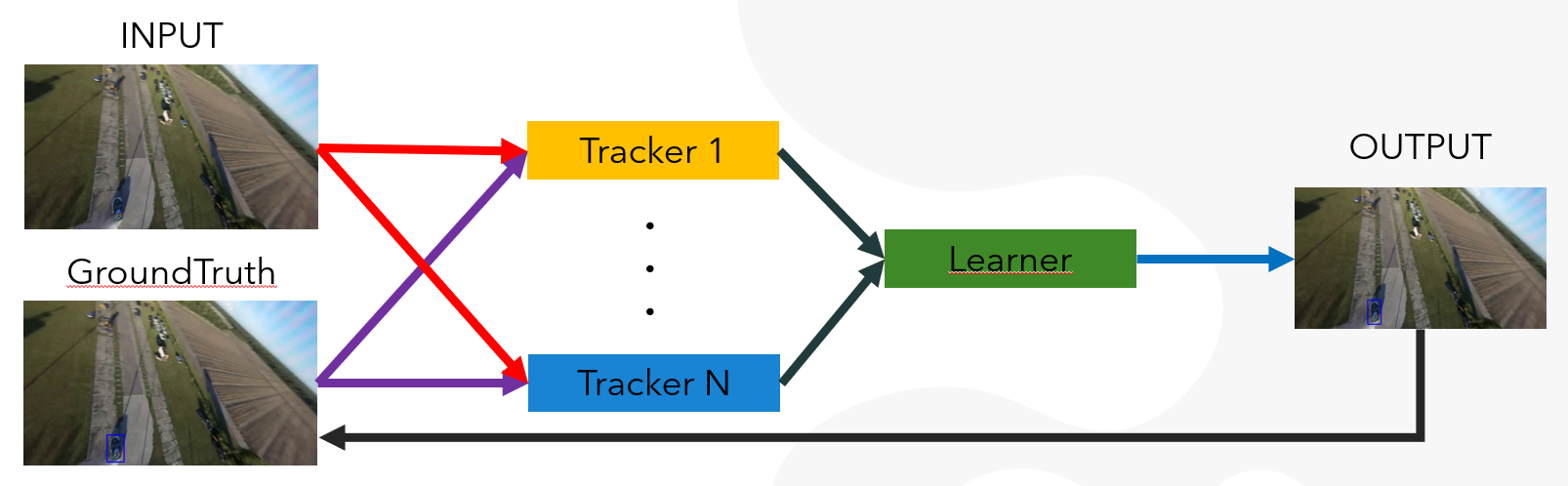}
	\caption{The entire proposed pipeline: both the input and the first target are passed to the $N$ trackers system, which predict a confidence score to be entered by the pre-trained learner. The learner will decide which bounding box to use based on the result of the score classification.}
	\label{lsdoltts}
\end{figure}
\begin{figure}[ht]
	\centering
		\includegraphics[scale=.50]{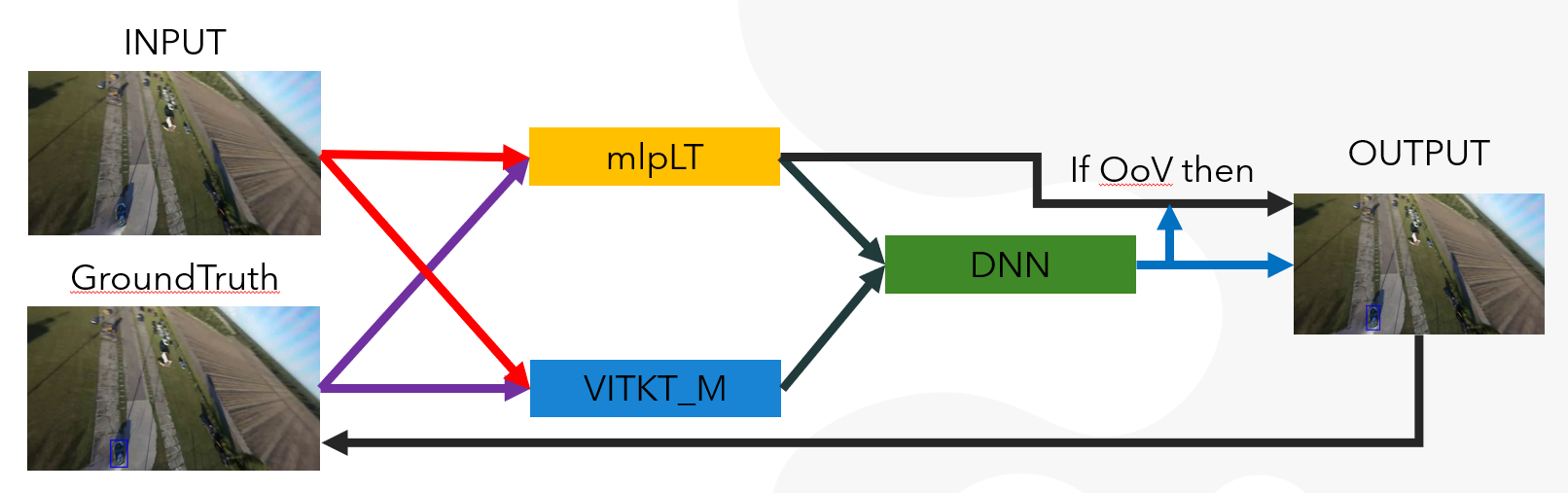}
	\caption{In the specialized model, two macro trackers are considered, mlpLT and VITKT\_M, and a DNN as a learner, trained on LTB-50 produced scores. In accordance with VOT standards, OoV should not be reported if predicted, so if there is no visibility detected, mlpLT is chosen, so that no priori knowledge is given on VOT-LT2022.}
	\label{lsdoltts2}
\end{figure}

\subsection{Proposed model}

In order to learn the behavior curve of the various trackers, and understand how their performance changes based on the output they predict and the groundtruth, the model we propose aims to train, on the scores predicted by them on each frame of every sequence, a ML algorithm. Regardless of the approach chosen, supervised or not, the goal is to decide which one among the $N$ trackers is actually the best choice, and to do this each training frame has to be noted with a corresponding classification label, where the classes are the indices of the trackers considered plus a label for the OoV, \textit{i.e.}, where the target is not clearly visible to any of the trackers. The ML model will then have $N$ scores of input (one for each tracker) and $N + 1$ output classes. An illustration of this process is given in Fig. \ref{lsdoltts}.
From the point of view of the trackers to be merged within the system, it was chosen to use the two best trackers currently known in the VOT field: mlpLT and VITKT\_M. In turn, the two trackers are composed of two sub-trackers, as already described in \textbf{Trackers Fusion Strategy}, so the system consists of a tournament of 4 trackers, carried out in two matches: in the first the results of the sub-trackers are verified, while in the second, which we could define as a final match, the scores produced go into input to the learning algorithm. The final predicted class will be the index of the tracker to use. The model chosen for training on LTB-50 and testing on VOT-LT 2022 is a Deep Neural Network (DNN). Specifically, the network is composed of two hidden layers: based on the dimensionality of the system, composed of an input of two scores, it was decided to place 3 hidden neurons at the first layer and 2 at the second layer, before arriving at the single output neuron. The last neuron can return 3 different states, \textit{i.e.}, choose the bounding box of mlpLT, the bounding box of VITKT\_M, or report an OoV. The DNN has been trained using the Limited-Memory BFGS optimization on a maximum number of iterations equal to $5000$. For DNN supervision, before the training phase, a standard transformation was applied to the data with respect to mean and variance. Likewise, the same transformation is applied before evaluating the data being tested, compared to the mean and variance model calculated on the training. According to the VOT evaluation rules, the OoV does not have to be reported, but a bounding box can be returned in any case, if foretold. This involves the arbitrary choice of an outcome, which heuristically falls between the two trackers. Being VITKT\_M the winner of the VOT-LT 2022 challenge, use it to give these bounding boxes would mean cheating on results. For this reason, the resulting winning model is used on the training dataset, so that there is no prior knowledge about the test set. Then, in case of OoV prediction, the result of mlpLT will be used on the VOT-LT 2022 test set. A visual representation of this architecture is contained in Fig. \ref{lsdoltts2}. The tracker proposed in the experiments turns out to be a cross between an online tracker, which is mlpLT, and an offline tracker, that is, VITKT\_M. The final part, the Machine Learning-based module, works on the current frame and can be considered online.

\subsection{Rationale behind DNN}

To choose the right node configuration for the model that uses DNN as a learner, it was decided to appeal to the Vapnik-Chervonenkis Dimension applied to Multi-Layer Perceptron (MLP) with ReLU as activation function. The VC dimension is the maximum number $n$ of scattered points from a binary classifier such that they can assume correct labeling. Considering the VOT protocol without out of view, the algorithm returns only two classes, namely the outputs of the first or second tracker, so it can be considered binary. According to Remark 9 in \cite{vc1}, a strict size $\Theta_1$ limit is imposed in the above case.
\begin{equation}
cWLlog(\frac{W}{L}) \leq VC \leq CWLlogW
\label{eqn21}
\end{equation}
From eqn. \ref{eqn21} it follows that there must be two constants $c$ and $C$ such that the strict limit condition is fulfilled, where $W$ and $L$ are the weights and layers of the network, respectively, and VC its Vapnik-Chervonenkis dimension. For simplicity, we can consider $c = C$, since at the right member the amount $log(W)$ is certainly greater than $log(\frac{W}{L})$. Having now to find a value of VC and C that solve the inequality according to our problem, a second condition is needed: we have referred to the sample-complexity bounds, which consists of another strict limit $\Theta_2$.
\begin{equation}
a\frac{VC+log(\frac{1}{\rho})}{\sigma} \leq N \leq b\frac{VC+log(\frac{1}{\rho})}{\sigma}
\label{eqn22}
\end{equation}
Similarly to the penultimate equation, the eqn. \ref{eqn22} admits as true the strict limit condition of $N$ number of training patterns if there are two constants $a$ and $b$, with $\rho$ the failure probability (which we considered to be the fraction of misclassified patterns on the test) and $\sigma$ the learning error (which was considered from the last value obtained on the training loss). This $\Theta_2$ comes from the combination of the upper bound treated in \cite{vc2} and the lower bound in \cite{vc3}.
Now, having a system of 4 inequalities in 4 variables, it was decided to impose for simplicity the constants $C = 1$ and $a = 1$. In this way, fixing the number of training patterns at $N = 215294$ (number of frames of LTB-50 \citet{lukezic2021}), with 3 nodes at the first hidden layer and 2 at the second, it gives $W = 14$ and $L = 4$. In addition, we obtained a $\rho = 0.45$ and $\sigma = 0.80$. The system now allows a solution for $b = \frac{4359687}{3682}$ and $VC = \frac{3682}{25}$. Since the solution to the system for the chosen constants exists, it can be asserted that the chosen configuration falls within the eligible configurations. The solution does not imply that the system does not overfit: this depends on the proportion between the number of training patterns and dimensionality, but above all on the quality and statistical independence of the patterns.

\begin{table}[width=.9\linewidth,cols=4,pos=ht]
\caption{Results and comparison with accepted methods presented at VOT-LT2022, sorted by F1-Score}\label{tbl1}
\begin{tabular*}{\tblwidth}{@{} LLLL@{} }
\toprule
Method & Precision & Recall & F1-Score\\
\midrule
VITKT\_M & 0.629 & 0.604 (2°) & 0.617 \\
mixLT & 0.608 & 0.592 & 0.600 \\
HuntFormer & 0.586 & 0.610 & 0.598 \\
CoCoLoT & 0.591 & 0.577 & 0.584 \\
Proposed model (DNN) & 0.562 & 0.619 (1°) & 0.582 \\
mlpLT & 0.568 & 0.562 & 0.565 \\
Proposed model (FCM) & 0.538 & 0.593 (3°) & 0.564 \\
KeepTrack & 0.572 & 0.550 & 0.561 \\
D3SLT & 0.520 & 0.516 & 0.518 \\
SuperDiMP & 0.510 & 0.496 & 0.503 \\
\bottomrule
\end{tabular*}
\end{table}

\begin{table}[width=.9\linewidth,cols=4,pos=ht]
\caption{Results and comparison with accepted methods presented at VOT-LT2021, sorted by F1-Score}\label{tbl2}
\begin{tabular*}{\tblwidth}{@{} LLLL@{} }
\toprule
Method & Precision & Recall & F1-Score\\
\midrule
mlpLT & 0.741 & 0.729 (3°) & 0.735 \\
VITKT\_M & 0.728 & 0.719 & 0.724 \\
STARK\_LT & 0.721 & 0.725 & 0.723 \\
STARK\_RGBD\_LT & 0.719 & 0.724 & 0.721 \\
SLOT & 0.727 & 0.711 & 0.719 \\
Keep\_track\_lt & 0.725 & 0.700 & 0.712 \\
SuperD\_MU & 0.738 & 0.680 & 0.708 \\
RincTrack & 0.717 & 0.696 & 0.707 \\
Proposed model (FCM) & 0.658 & 0.738 (1°) & 0.696 \\
LT\_DSE & 0.715 & 0.677 & 0.695 \\
Proposed model (DNN) & 0.653 & 0.732 (2°) & 0.690 \\
LTMU\_B & 0.698 & 0.680 & 0.689 \\
SuperDiMP & 0.675 & 0.660 & 0.667 \\
SiamRCNN & 0.654 & 0.673 & 0.664 \\
Sion\_LT & 0.640 & 0.456 & 0.533 \\
TDIOT & 0.496 & 0.478 & 0.487 \\
\bottomrule
\end{tabular*}
\end{table}

\begin{table}[width=.9\linewidth,cols=4,pos=ht]
\caption{Out of view detection skill for best models on VOT-LT 2022 and LTB-50. $OoV_P$ stands for out-of-view predictions while $OoV_G$ stands for out-of-view total number.}\label{tbl3}
\begin{tabular*}{\tblwidth}{@{} LLLL@{} }
\toprule
Method & $OoV_P$ & $OoV_G$ & Test\\
\midrule
Proposed model (DNN) & 10155 & 16733 & VOT-LT2022 \\
Proposed model (FCM) & 18119 & 27310 & LTB-50 \\

\bottomrule
\end{tabular*}
\end{table}

\subsection{Implementation}
Details on individual implementations and libraries to be installed from Github can be found in the mlpLT and VITKT\_M reference papers, or on the official VOT challenge website. The learners were implemented using the Python scikit-learn and fuzzy-c-means libraries. The official repository of this work is available at the link \url{https://github.com/knapsack96/lsdoltts}. The environment used for the experiments is Kaggle, a Google cloud tool that provides free computational and storaging resources.
\begin{figure}[ht]
	\centering
		\includegraphics[scale=.50]{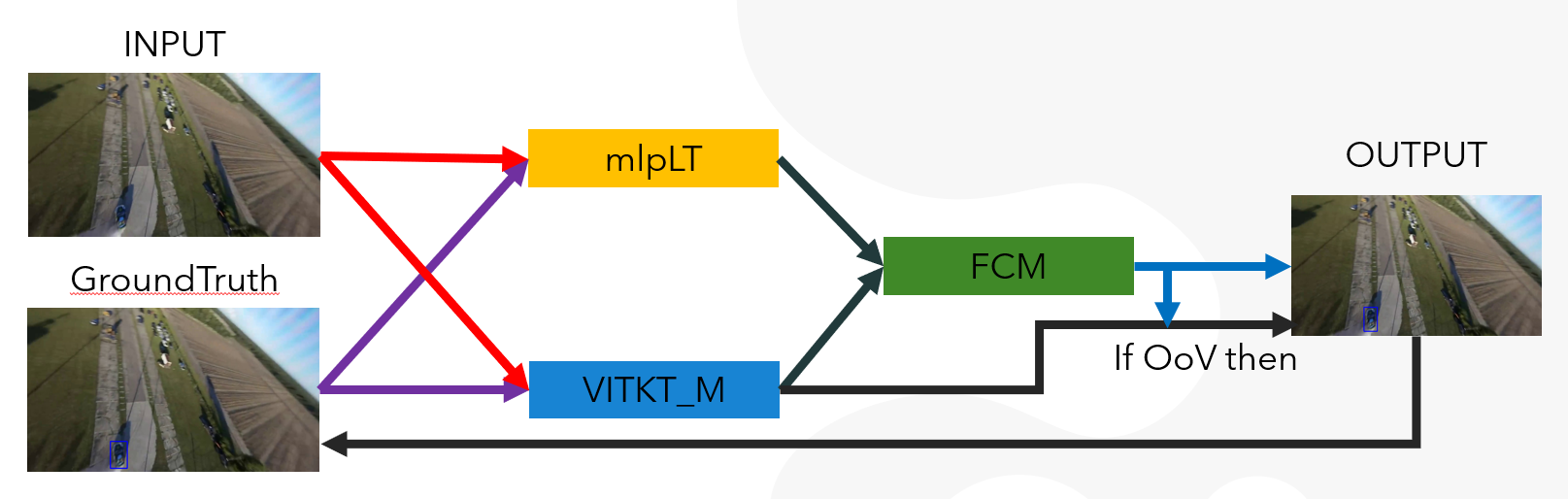}
	\caption{As already shown in Fig. \ref{lsdoltts2}, the main trackers are mlpLT and VITKT\_M, while the learner is unsupervised. In this case, it consists of a Fuzzy C-Means. The OoV case is addressed by choosing the VITKT\_M outcome.}
	\label{lsdoltts1}
\end{figure}

\begin{figure}[ht]
    \centering
        \includegraphics[scale=.90]{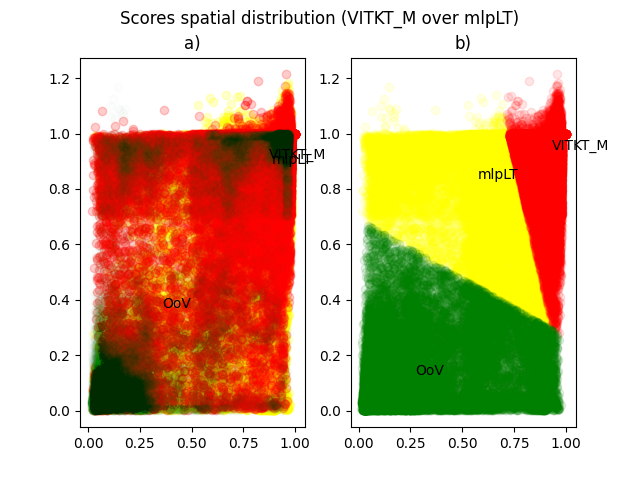}
		
	\caption{Spatial distribution of scores: a) The scores of mlpLT and VITKT\_M mapped into a 2D space, highlighting in yellow the points that indicate a situation in which mlpLT was better, in red for VITKT\_M and finally in green for an OoV. The points consist of the groundtruth calculated on the LTB-50 dataset. As you can see, there are areas of high density subject to interference among classes; b) In the same space of the scores defined in a), the points are clustered according to the fuzzy criterion and then a hard assignment is adopted, outlining three well-defined areas of belonging.}
	\label{fcm}
\end{figure}

\section{Results and discussion}
In this section, results obtained from the experiments are discussed, placing the proposed method within the ranking of the various baselines. In addition, some ablation studies are proposed to verify the consistency of the results in different set-ups. 

\subsection{Ablation study}

To verify that changing the type of learner or type of data on which to train the results obtained remained consistent, some changes were made.
It was decided in a second phase to use an unsupervised learning-based learner, in particular referring to fuzzy logic clustering. The main reason for this choice lies in the fact that the distributions of the scores of the trackers are by nature overlapping: as already discussed in the section \textbf{Trackers complementarity}, despite having monotony between two scores, it is not said that there is monotony on the IoU of the relative bounding boxes. Fuzzy logic allows us to assign a value of belonging of a point to various sets, or to blur the classification task. This ability allows to reduce, where possible, the noise in the areas of overlap of the scores. In the case of fuzzy c-means (FCM) as learner, no actual supervised annotation was made, but after the clustering phase the points were assigned to the cluster with the highest membership value, in order to maximize the accuracy in terms of classification. The FCM parameters concern a number of cluster equal to the number of trackers plus $1$, so $3$ in our case, and a degree of fuzzy overlap equal to $2$ (the least possible for FCM).
Other ablations refer to the inversion of datasets for training and testing. Two experiments were then carried out using the two learners DNN and FCM, setting the VOT-LT2022 dataset as the training set and LTB-50 as the test set. In this case, if an OoV is detected, the process mentioned in \textbf{Proposed model} is inverted, so the VITKT\_M result is chosen in place of mlpLT. The other version of the pipeline (using the FCM) is visible in Fig. \ref{lsdoltts1}.
Results of both methods with DNN and FCM learners are displayed in Table \ref{tbl1} when training on LTB-50 and testing on VOT-LT 2022 and viceversa on Table \ref{tbl2}. Also in the inversion of the two datasets the resolution of the system of inequalities defined in \textbf{Rationale behind DNN} was tested, where this time $N = 168282$ (number of frames of VOT-LT2022 \citet{kristan2023}), $\rho = 0.52$ and $\sigma = 0.81$. Keeping the same constants as the first experiment, we admit a solution for the system with $b = \frac{4359687}{3682}$ and $VC = \frac{3682}{25}$.

\subsection{Discussion}

Looking at the results obtained in Table \ref{tbl1}, the proposed method is to be with both DNN and FCM learners among the first three places for recall (respectively 1st and 3rd), while in second place is VITKT\_M. The same situation recurs in Table \ref{tbl2}, where even the method takes 1st and 2nd place in the recall, reversing the learners, or this time FCM turns out to have the best recall on LTB-50. In third place is mlpLT. Considering the F1-Score, it is interesting to note that in Table \ref{tbl1} the method presents with DNN learner a higher value of mlpLT, and at worst, using FCM, a higher value than KeepTrack. The best F1-Score (obtained with DNN), is in fifth place on VOT-LT 2022, after CoCoLoT, extension of mlpLT. Similarly, in Table \ref{tbl2}, it is important to note how the method with FCM learner exceeds LT\_DSE (the winner of the 2019 and 2020 editions) confirming the superiority of the merged approach. The method, for this metric, ranks 9th on the LTB-50. 

\begin{figure}[ht]
	\centering
		\includegraphics[scale=.33]{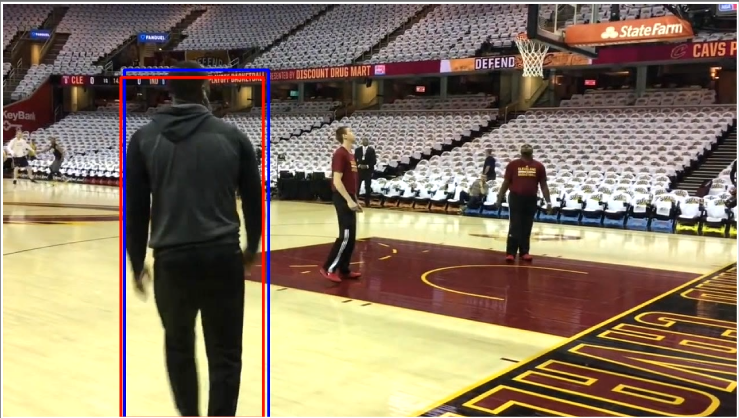}
  \includegraphics[scale=.33]{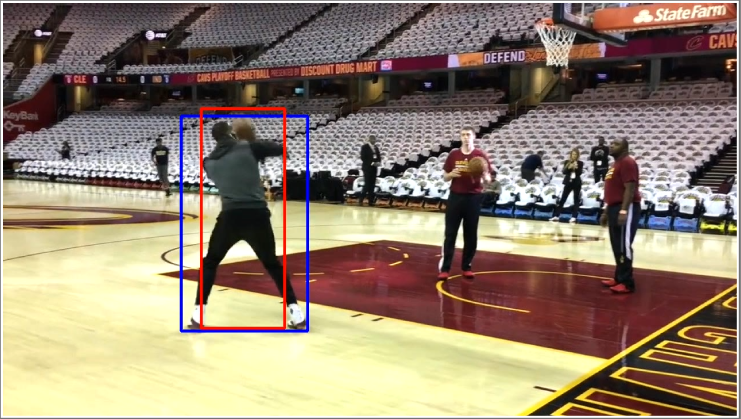}
  \includegraphics[scale=.33]{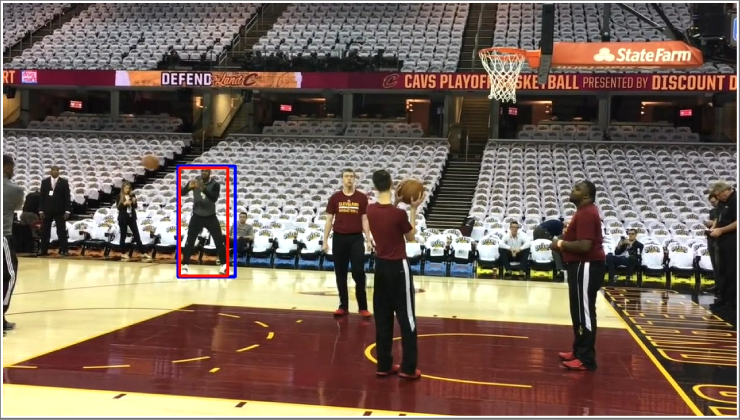}
  
	\caption{Visual results of the model with FCM learner trained on VOT-LT2022 on a sequence of the LTB-50 dataset; In red are represented the bounding boxes of groundtruth while in blue those predicted.}
	\label{basket}
\end{figure}

\begin{figure}[ht]
	\centering
		\includegraphics[scale=.33]{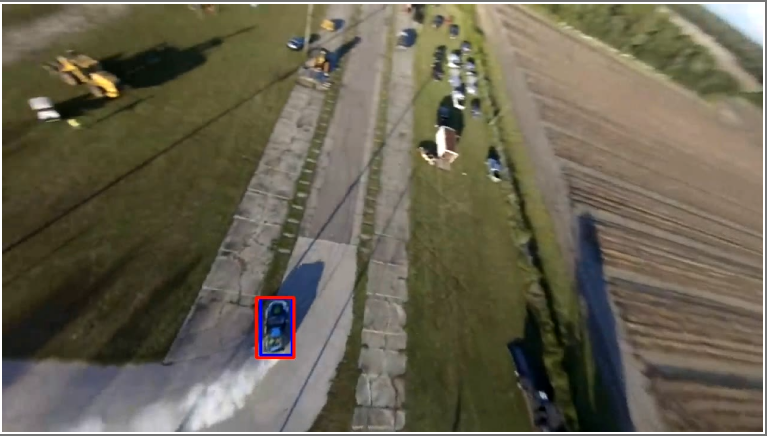}
  \includegraphics[scale=.33]{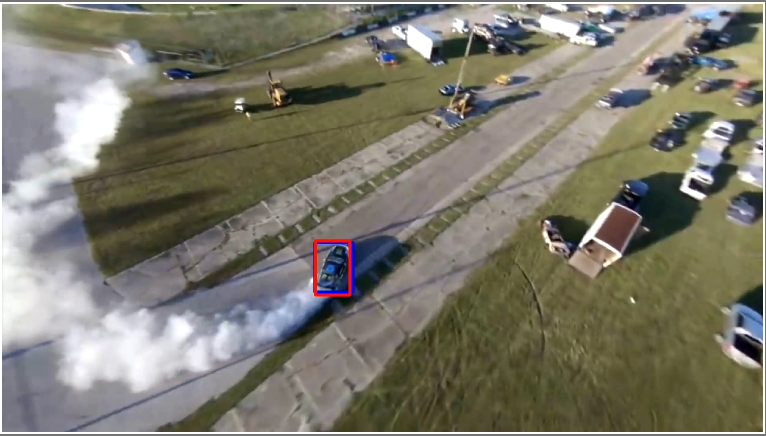}
  \includegraphics[scale=.33]{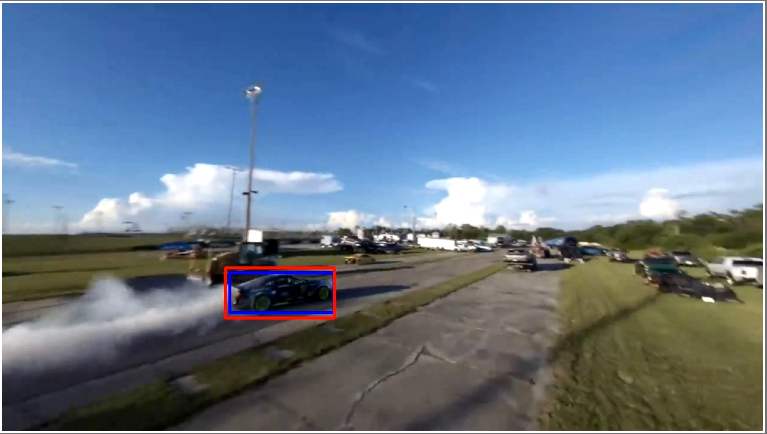}
	\caption{Visual results of the model with DNN learner trained on LTB-50 on a sequence of the VOT-LT2022 dataset; In red are represented the bounding boxes of groundtruth while in blue those predicted.}
	\label{car}
\end{figure}

Different is the speech of precision, which turns out to be much lower than the recall, lowering the average of the F1-Score. In Table \ref{tbl1} it is around 7th place, while in Table \ref{tbl2} it goes to 11th. The context is defined by the trackers accepted and submitted to the editions. The result found on the LTB-50 dataset with FCM learner are shown in Fig. \ref{fcm}, where the spatial distribution of the scores has as coordinates the confidence of mlpLT on the abscissa axis and that of VITKT\_M on the ordinates. As described in the figure, in part \textbf{a)} the groundtruth distribution contains a high interference density between classes, while in \textbf{b)} it is demonstrated how the application of FCM manages to balance the three areas to which it belongs. It is clear how the application of an ML method may be necessary to reduce the overlap rate that resides between the scores of the algorithms, where the target is fully visible (yellow and red areas) and not visible (green zone). The introduction of a learning phase of the scores also allows a better control over the detection of the OoVs, as shown in Table \ref{tbl3}, where both the best experiments were evaluated on the basis of the number of OoV predicted well on the total. In both cases, the percentage of true positive OoVs is approximately 66\%. The numerical results can be better interpreted by looking at the Fig. \ref{basket} and \ref{car}, where 4 frames of two sequences of the two experiments are shown, where the groundtruth bounding boxes are drawn in red and the aforementioned ones are drawn in blue. In Fig. \ref{basket} the model with FCM learner is applied, while in Fig. \ref{car} the one with DNN learner; In both figures you can see the decent quality of the tracking system compared to the groundtruth. Having obtained the above results, it can be confirmed the conjecture that the proposed method enjoys two important properties: the \textit{model-independence}, or the ability to improve results by combining different trackers independently of the type of learning chosen and the \textit{data-independence}, or the ability to keep the previous property unchanged while changing the training and test data.

\begin{figure}[ht]
	\centering
		\includegraphics[scale=.55]{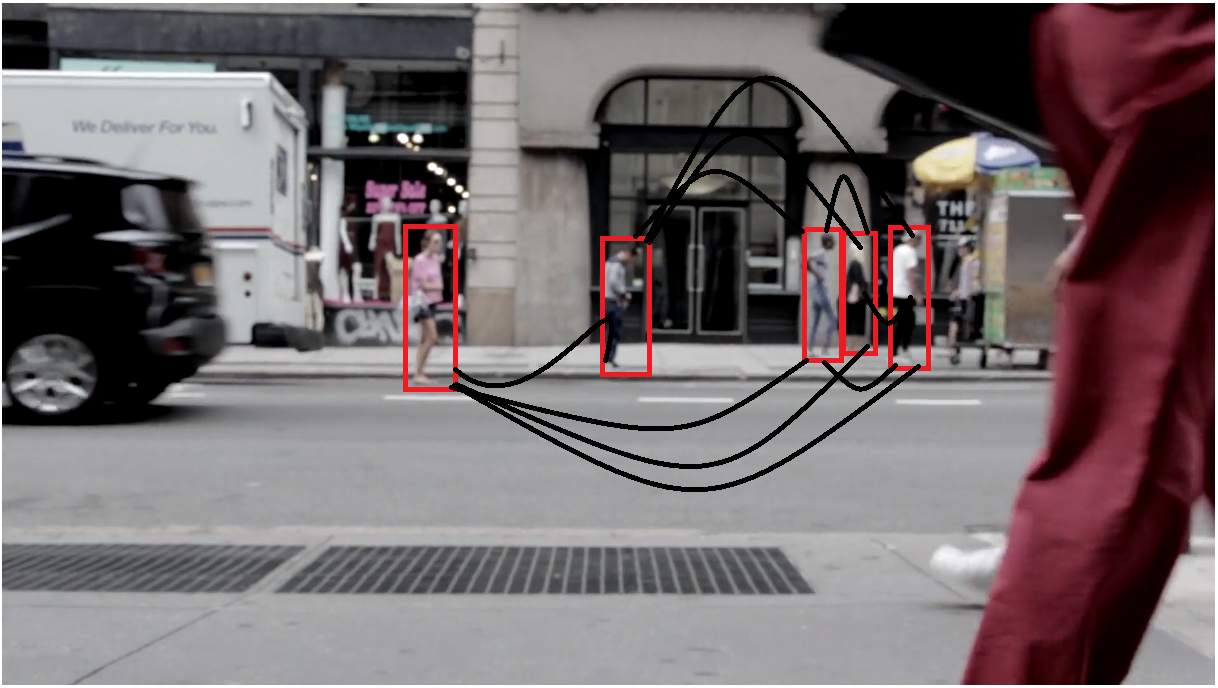}
	\caption{Graph encoding of 1st frame targets, where red bounding boxes mean visible nodes. The image comes from the following free and sharable video \url{https://www.pexels.com/video/people-walking-by-on-a-sidewalk-854100/}}
	\label{mot1}
\end{figure}
\begin{figure}[ht]
	\centering
		\includegraphics[scale=.55]{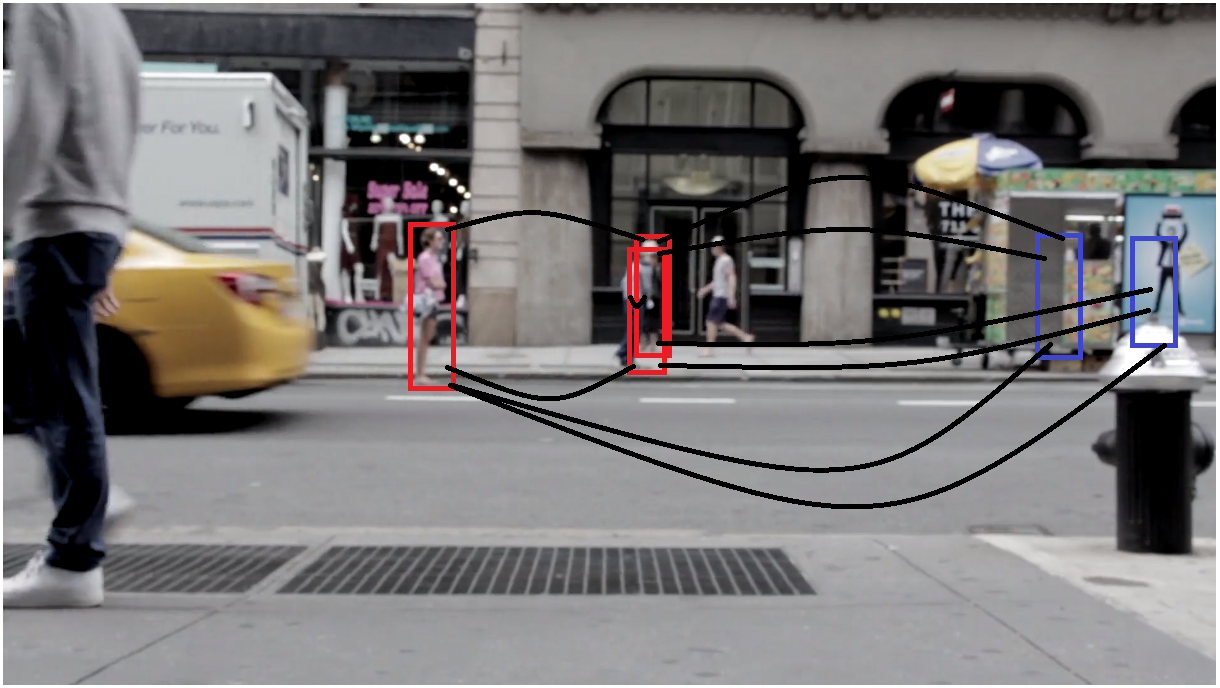}
	\caption{Graph encoding of n-th frame targets, where red bounding boxes mean visible nodes and blue bounding boxes mean out of view nodes. The image comes from the following free and sharable video \url{https://www.pexels.com/video/people-walking-by-on-a-sidewalk-854100/}}
	\label{mot2}
\end{figure}

\subsection{Extension to multi-object tracking}

Similarly to single-object tracking, it would not be absurd to think that the approach presented could be extended to multiple targets. In the present case, the trackers to be taken into consideration for the merger would be Multiple Object Trackers (MOT), for example the best at the state of the art, while the learner, being several targets having their own confidence score, their own presence / absence from the scene and their own identifier, could be structured for instance as GNN. Input graphs would admit individual targets as nodes and score functions as edges: an example would be an all-connected graph in which each edge is the weighted average between the two scores compared to the distance in pixels. 
In addition, nodes might have attributes such as their own unique identifier and their own \textit{bit} representing the binary state of presence or absence (OoV). In this sense, the GNN would perform a graph classification task among several MOTs, choosing the best graph to represent the output. When a node is not visible, its distance in pixel is ignored on the edge weight, as it could be unknown. In Fig. \ref{mot1} a frame with the abovementioned graph encoding is shown, with black circles as nodes and black lines as edges. In Fig. \ref{mot2} a consecutive frame of the same sequence is shown, with one of the target not visible anymore: the OoV has been depicted as a white circle in an arbitrary position, still connected to the other nodes. In an advanced, multi-target-oriented version of the proposed method, this idea could be considered to test the conjecture and thus the properties of model-independence and data-independence.

\section{Conclusion}
A new tracker fusion approach to the problem of long-term single-object tracking has been presented. In the manuscript, the generalization of the number of tracker components of the system, greater than the 2 usually used by most models, the ability to improve an ensemble of trackers by adding a final learning phase on the scores produced and the introduction of a classification of non-visible or OoV targets was discussed. A conjecture on the new paradigm has been formulated, theorizing the property of model-independence and data-independence, and an extension of the approach to multi-object tracking has been introduced. The model's results improved on two state-of-the-art benchmarks in terms of recall, and ranked among the top in terms of F1-score. 

\printcredits
\bibliographystyle{cas-model2-names}

\bibliography{cas-refs}

\vskip3pt
\bio{figs/enzo}
Vincenzo Mariano Scarrica is PhD student at the National PhD Program in AI – Agrifood and Environment, University of Naples Federico II. He previously obtained a master's degree in "Machine Learning and Big Data" at the DiST (Department of Science and Technology) of the University of Naples Parthenope, with which he actively collaborates. During his training he has dealt with various issues related to AI, in particular in the area of Computer Vision.
\endbio

\hfill  \break
\hfill  \break
\hfill  \break
\bio{figs/profstaiano}
Antonino Staiano is Associate Professor at the Department of Science and Technology of the University of Naples Parthenope where he is the owner of the Natural Language Processing and Artificial Intelligence courses of the Master's Degree Course in Machine Learning and Big Data and is the scientific director of the interdisciplinary laboratory Neptun- AI. His research activity is focused on the development of Machine Learning techniques applied to various fields such as Astrophysics, Bioinformatics and the Environment with particular attention to the themes of the Sea. He is Director of the Parthenope Research Unit of CINI.
\endbio

\end{document}